\documentclass[a4paper]{article}
\usepackage{geometry}
\usepackage{indentfirst}
\usepackage[pdftex]{graphicx}
\usepackage{amsmath}
\usepackage{subfigure}
\usepackage{cite}
\usepackage{authblk}
\usepackage{titlesec}
\usepackage[bookmarks=true,colorlinks,linkcolor=black,citecolor=black,urlcolor=black]{hyperref}
\usepackage{natbib}

\setcitestyle{authoryear,open={(},close={)}}

\graphicspath{{Graphics/}}
\renewcommand\footnotemark{}

\title{\textbf{Impedance Control of a Cable-Driven SEA \\ with Mixed $H_2/H_\infty$ Synthesis}
	\thanks{Manuscript received on November 24, 2016; Revised on February 13, 2017; Accepted on February 23, 2017.}
	\thanks{This paper has been accepted and published in the journal \textbf{Assembly Automation}. Citation:
		Ningbo Yu, Wulin Zou, (2017) ``Impedance control of a cable-driven SEA with mixed $H_2/H_\infty$ synthesis'', Assembly Automation, Vol. 37, Issue: 3, pp.296-303, https://doi.org/10.1108/AA-11-2016-150.}
	\thanks{Corresponding author: Ningbo Yu, Email: nyu@nankai.edu.cn.}
}
\author[1,2]{Ningbo Yu*}
\author[1,2]{Wulin Zou}
\affil[1]{\small Institute of Robotics and Automatic Information Systems, Nankai University}
\affil[2]{\small Tianjin Key Laboratory of Intelligent Robotics, Nankai University\vspace{-11ex}}
\date{}

\begin{document}
	
\maketitle

\section*{Abstract}

\textbf{\textit{Purpose}}: This paper presents an impedance control method with mixed $H_2/H_\infty$ synthesis and relaxed passivity for a cable-driven series elastic actuator to be applied for physical human-robot interaction.

\textbf{\textit{Design/methodology/approach}}: To shape the system's impedance to match a desired dynamic model, the impedance control problem was reformulated into an impedance matching structure. The desired competing performance requirements as well as constraints from the physical system can be characterized with weighting functions for respective signals. Considering the frequency properties of human movements, the passivity constraint for stable human-robot interaction, which is required on the entire frequency spectrum and may bring conservative solutions, has been relaxed in such a way that it only restrains the low frequency band. Thus, impedance control became a mixed $H_2/H_\infty$ synthesis problem, and a dynamic output feedback controller can be obtained.

\textbf{\textit{Findings}}: The proposed impedance control strategy has been tested for various desired impedance with both simulation and experiments on the cable-driven series elastic actuator platform. The actual interaction torque tracked well the desired torque within the desired norm bounds, and the control input was regulated below the motor velocity limit. The closed loop system can guarantee relaxed passivity at low frequency. Both simulation and experimental results have validated the feasibility and efficacy of the proposed method.

\textbf{\textit{Originality/value}}: This impedance control strategy with mixed $H_2/H_\infty$ synthesis and relaxed passivity provides a novel, effective and less conservative method for physical human-robot interaction control.

\section*{Keywords}
Impedance control, Human-robot interaction, Mixed $H_2/H_\infty$ synthesis, Relaxed passivity, Series elastic actuator

\section*{Paper Type}
Research paper

\section{Introduction}

For many applications where robots directly interact with human body or unstructured external environment, safety and compliance are of significant importance. An effective solution, initially proposed by~\citep{Pratt1995}, is the series elastic actuator (SEA), in which elasticity is intentionally introduced in series between the motor and end-effector. This structure became increasingly popular during the last decades~\citep{Robinson1999,Yoo2015,ZouWulin2016IROS} for a number of advantages, such as great shock tolerance, safety, energy storage, accurate and smooth force output, etc. To detach the actuation motor from the end-effector and transmit torque to a distant place, the cable-driven SEA structure has been introduced into physical human-robot interaction (pHRI) systems. Cable-driven SEAs are now widely used for various applications, e.g., for lower limb exoskeletons~\citep{Veneman2006}, for arm and wrist rehabilitation robots~\citep{Oblak2010}, for elbow joint exoskeletons~\citep{Lu2015}, for magnetic resonance imaging (MRI) compatible robots~\citep{Senturk2016}, etc.

Impedance control, first described by Hogan in~\citep{Hogan1985}, has been widely adopted in the control of robotic systems to regulate the compliance of interaction or contact tasks. The impedance represents the dynamic relationship between the motion and interaction force or torque. Thus, it is natural to look for a solution to manage the impedance with force/torque or position control in the existing control literature. A variety of such control architectures have been proposed for different SEAs. For example, in~\citep{Oblak2010,Sergi2015,Vallery2008}, pure proportional-integral-differential (PID) torque control was applied for SEA impedance control. In~\citep{Kong2009,Mehling2015IROS,Yu2015}, disturbance observer (DOB) based torque control methods were used to improve the impedance rendering accuracy and robustness. The aforementioned impedance control approaches were mostly implemented with the cascaded impedance-torque or impedance-torque-velocity control structure, wrapping impedance control around an inner torque or position control loop. The gains of the inner controller usually should be carefully selected to guarantee stability and obtain satisfying tracking performance. However, the controller performance usually could not be guaranteed to a certain boundary or be evaluated with quantifiable metrics.

It is beneficial to obtain guaranteed performance and to bring ease for controller design. In~\citep{Mehling2014}, a model matching framework based on $H_\infty$ control was constructed to directly control the impedance. Motivated by this approach, the impedance control problem of the cable-driven SEA will be reformulated to a mixed $H_2/H_\infty$ synthesis problem to improve the performance in this paper. In this framework, a well performing controller can be designed straightforwardly, and it is not necessary any more to design a torque sub-controller in advance. The maximum impedance rendering error can be guaranteed to an $H_\infty$ norm bound, and the maximum control effort can be bounded by an $H_2$ norm. The quantified norm bounds are closely related to system's performance requirements and physical constraints. This framework also makes it possible and easier to design a complex and arbitrary higher-order dynamic controller for high-precision structured impedance control. Further, it provides possibilities for multi-objective optimization, making those competing performances, e.g., tracking error, energy consumption, disturbance rejection and noise rejection, adjustable by weighting functions. Mixed $H_2/H_\infty$ synthesis can usually be solved by many ways, e.g., the linear matrix inequalities, the Riccati equations and the MATLAB Robust Control Toolbox, which will be a great advance from existing cascaded impedance control approaches.

SEAs should also maintain robust stability when interacting with the unstructured outside world. Colgate analyzed dynamically interacting systems in detail, and took passivity to guarantee interaction stability~\citep{Colgate1988}. Necessary and sufficient conditions related to the impedance of the driving point were derived for passive interaction. Nevertheless, those conditions usually bring conservative solutions for pHRI, which mostly falls in the low frequency band.

In this paper, a mixed $H_2/H_\infty$ synthesis-based strategy has been proposed for the impedance control of a cable-driven SEA platform for pHRI applications. The passivity constraints, which were usually conservative, have been relaxed to achieve guaranteed stable interaction only at the low frequency band. Our proposed method has shown good performance in practice and can be applied in some potential applications, such as the rehabilitation and assistive robots~\citep{Yu2014Robio,Yu2016AAS}, and the bio-inspired robots~\citep{Qiao2016AA}.

The paper is organized as follows. Section~\ref{section2} describes the cable-driven SEA platform and model, and the reformulation of the impedance control problem. Details for the $H_2/H_\infty$ controller synthesis are provided in Section~\ref{section3}. Simulations, experiments and results are presented in Section~\ref{section4}. Finally, Section~\ref{section5} concludes the paper.

\section{The cable-driven series elastic actuator impedance control problem}
\label{section2}

\subsection{The cable-driven series elastic actuator platform}

\begin{figure}[!h]
	\centering
	\includegraphics[width=0.65\columnwidth]{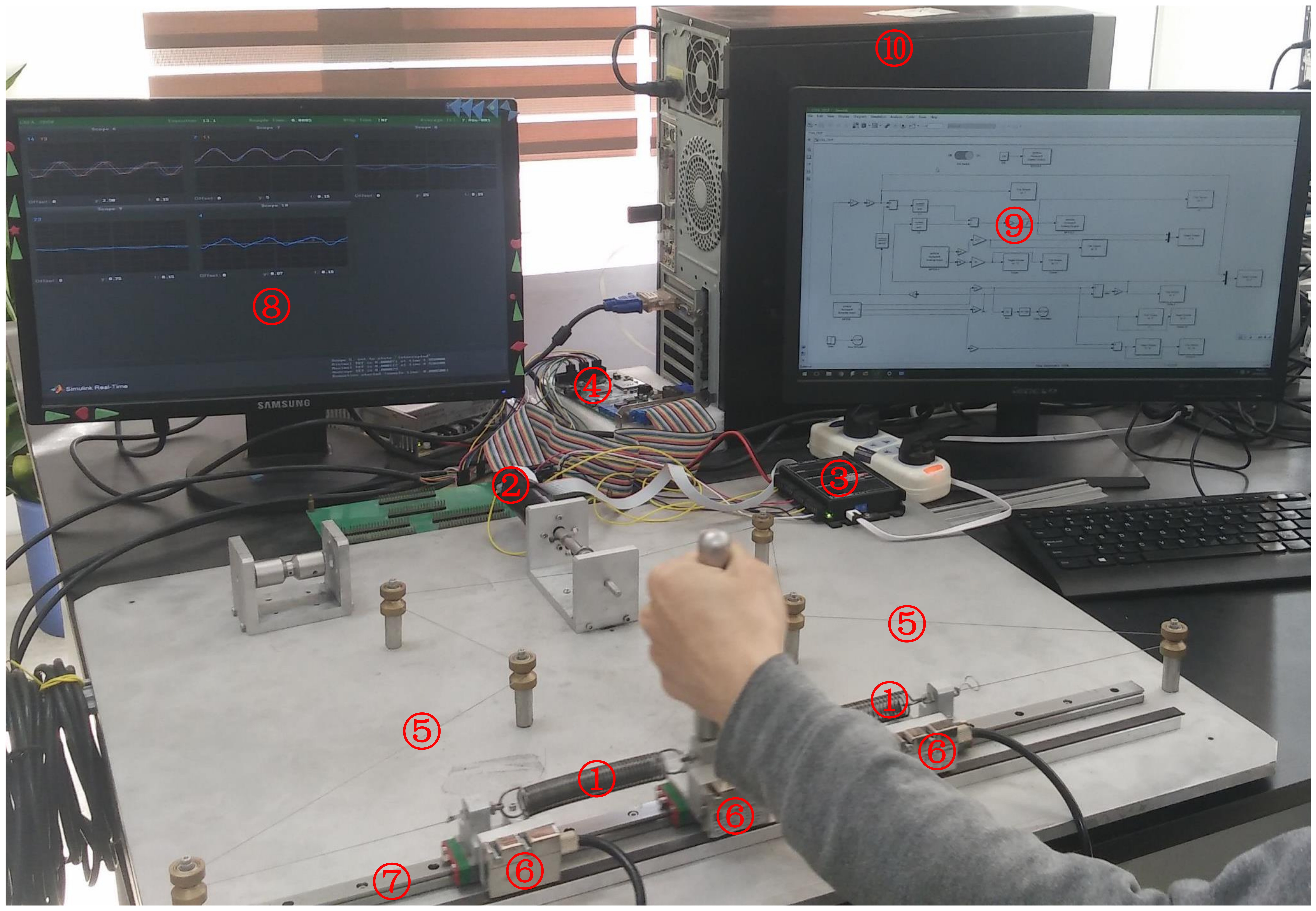}
	\caption{The cable-driven SEA platform. (1) the linear tension springs; (2) the Maxon DC motor; (3) the servo controller for velocity control; (4) the MF634 multifunction I/O card; (5) transmission cables; (6) magnetic linear encoders for displacement measurement; (7) the sliding guide; (8) scopes on the target computer for signal display; (9) the host computer; (10) the target computer running a real-time kernel.}
	\label{fig_prototype}
\end{figure}

A realized cable-driven SEA platform for pHRI is illustrated in Fig.~\ref{fig_prototype}. The handle can slide along the linear guide, as a result of the interaction between the human hand and the cable-driven SEA. Data acquisition and control are realized by a host computer running MATLAB/Simulink, a target computer running the real-time kernel, and a Humusoft MF634 multifunction I/O card.

\begin{figure}[!h]
	\centering
	\includegraphics[width=0.65\columnwidth]{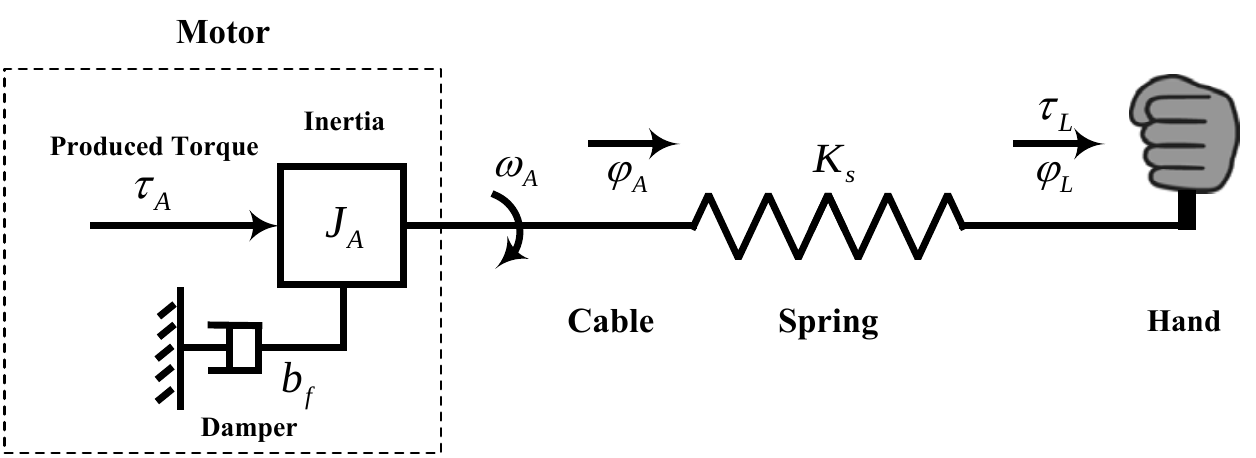}
	\caption{pHRI with the cable-driven SEA.}
	\label{fig_SEA}
\end{figure}

A human hand interacting with the cable-driven SEA platform is depicted equivalently in Fig.~\ref{fig_SEA}. The DC motor with inertia $J_A$ and damping $b_f$ produces the torque $\tau_A$. The motor velocity $\omega_A$ leads to the cable displacement $\varphi_A$. When interacting with the human hand, the resulted motion is denoted as $\varphi_L$ and the interaction torque acting on human hand is represented as $\tau_L$. The spring between the cable and hand has a stiffness of $K_s$. It is assumed that the cable has no elastic deformation and the inertia of the cable can be neglected.

\begin{figure}[!h]
	\centering
	\includegraphics[width=0.65\columnwidth]{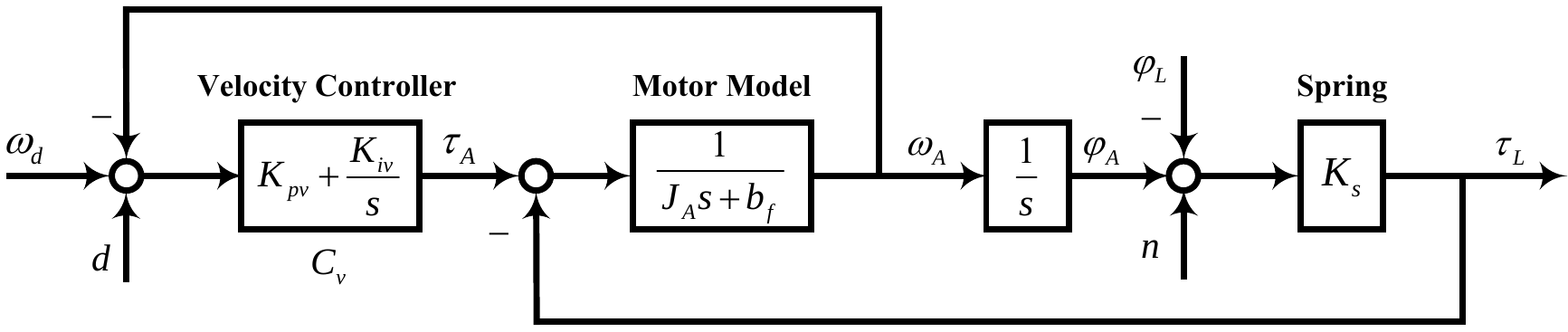}
	\caption{The linearized diagram of the cable-driven SEA with a velocity-controlled motor. $\omega_d$ is the desired velocity. $C_v$ is a PI motor velocity controller. $d$ and $n$ represent the input disturbance and sensor noise.}
	\label{fig_Velocity_Source_CSEA}
\end{figure}

As the interaction torque can be conveniently calculated by the measured spring displacement, the motor can be regarded as a velocity source to provide stable and accurate velocity for the cable-driven SEA~\citep{Wyeth2006}. The linearized diagram of the cable-driven SEA with a velocity-sourced motor model is shown in Fig.~\ref{fig_Velocity_Source_CSEA}. The dynamics from the four external inputs $\omega_d,\varphi_L,d,n$ to the output torque $\tau_L$ can be given as
\begin{equation}
{\tau _L}(s) = {G_1}(s)[{\omega _d}(s) + d(s)] + {G_2}(s)[{\varphi _L}(s) - n(s)],
\end{equation}
where 
\begin{eqnarray}
\displaystyle {G_1}(s) &=& \frac{{({K_{pv}}s + {K_{iv}}){K_s}}}{{{J_A}{s^3} + ({b_f} + {K_{pv}}){s^2} + ({K_{iv}} + {K_s})s}}, \\
\displaystyle {G_2}(s) &=& - \frac{{({J_A}{s^2} + ({b_f} + {K_{pv}})s + {K_{iv}}){K_s}}}{{{J_A}{s^2} + ({b_f} + {K_{pv}})s + {K_{iv}} + {K_s}}}.
\end{eqnarray}

\subsection{Reformulation of the impedance control problem}

Impedance is defined as the ratio of the torque applied to the mechanical system and the velocity of the interaction handle, such that:
\begin{equation}\label{eq_Z}
Z(s) = \frac{{{\tau _L}(s)}}{{ - {{\dot \varphi }_L}(s)}}.
\end{equation}
Then, the relationship between the torque and the handle motion is:
\begin{equation}\label{eq_PZ}
{P_Z}(s) = \frac{{{\tau _L}(s)}}{{ - {\varphi _L}(s)}}.
\end{equation}

\begin{figure}[!h]
	\centering
	\includegraphics[width=0.65\columnwidth]{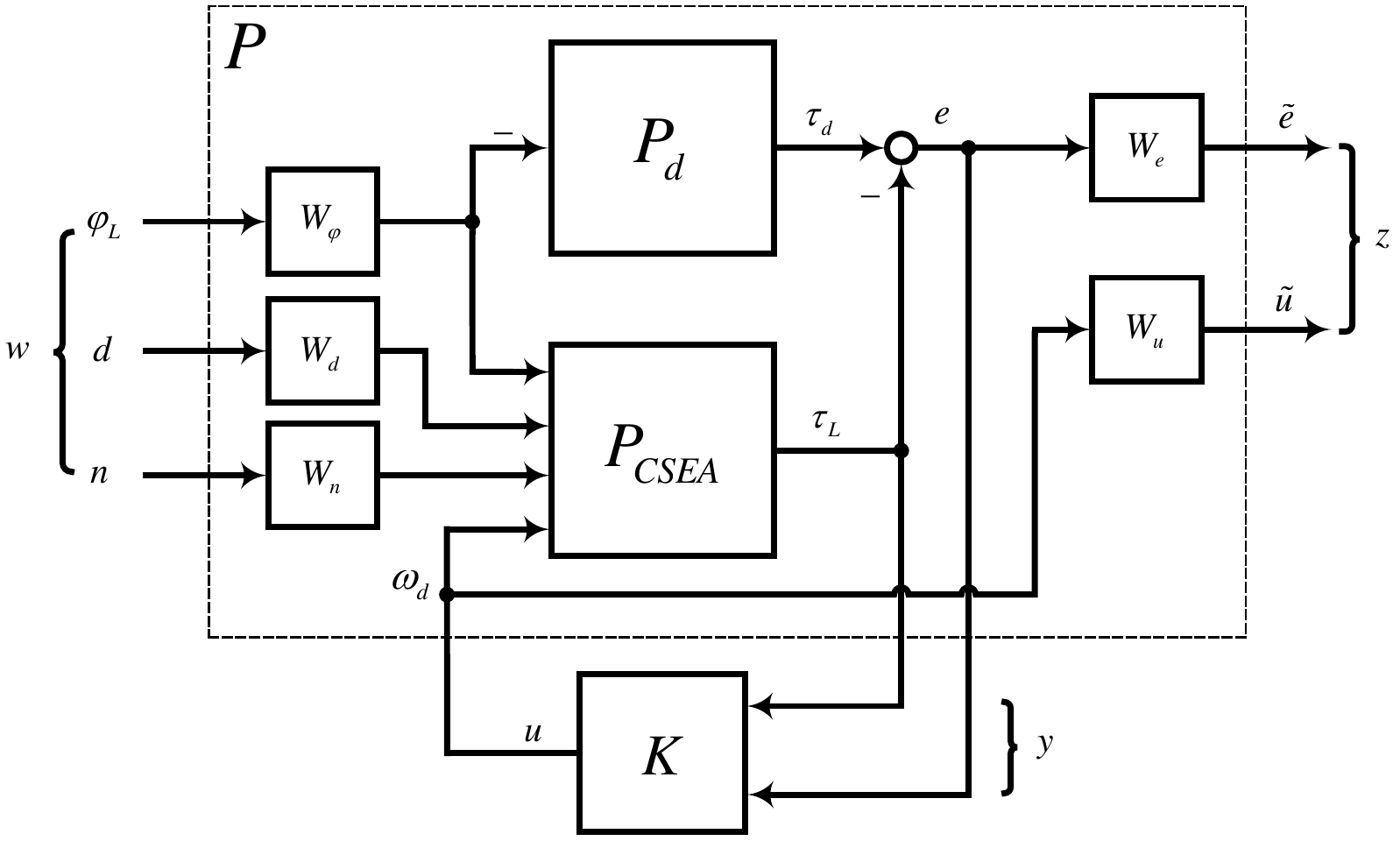}
	\caption{The impedance control strategy for the cable-driven SEA.}
	\label{fig_Impedance_Control_Synthesis}
\end{figure}

Impedance control is essentially to shape a given system's behavior $Z(s)$ to match a desired dynamic model $Z_d(s)$. It is equivalent to shaping the dynamics of $P_Z(s)$ to approximate the desired dynamic model $P_d(s)$. In consideration that the measured exogenous input is the human motion $\varphi_L$ instead of hand velocity in many cases, the impedance control structure illustrated in Fig.~\ref{fig_Impedance_Control_Synthesis} can be elaborated to achieve the goal of shaping the dynamics $P_Z(s)$. 

In this framework, $w=[\varphi_L,~d,~n]^{\rm T}$ is a vector of all external signals, including the reference signal, disturbance and noise. $z=[\tilde e,~\tilde{u}]^{\rm T}$ is a vector including the weighted tracking error and control input. $y=[\tau_L,~e]^{\rm T}$ can be directly calculated from sensor measurements. $u=\omega_d$ is the control signal going into the plant $P$. $\tau_d$ is the desired interaction torque. The block $W_*$ represents the weighting functions for each signal. $P_{CSEA}$ denotes the SEA model as shown in Fig.~\ref{fig_Velocity_Source_CSEA}. $K$ is the impedance controller, which is supposed to stabilize the system and regulate the impedance. The desired virtual impedance defined by equation~(\ref{eq_PZ}) takes the model encompassing virtual inertia $M_d$, virtual damping $B_d$ and virtual stiffness $K_d$, and thus:
\begin{equation}
{P_d}(s) =  {M_d}{s^2} + {B_d}s + {K_d}.
\end{equation}
Then, the impedance defined by equation (\ref{eq_Z}) is:
\begin{equation}
{Z_d}(s) =  {M_d}s + {B_d} + \frac{{{K_d}}}{s}.
\end{equation}

In this framework, it is first required to minimize or bound the impedance matching error $e$, or equivalently the weighted error $\tilde e$ in the presence of disturbance and noise. A second goal is to optimize the control input $u$ or equivalently the weighted control input $\tilde u$. This is a multi-objective control problem for the cable-driven SEA and the mixed $H_2/H_{\infty}$ control can be a viable method under this structure.

Considering the following transfer matrix:
\begin{equation}
\left[ {\begin{array}{*{20}{c}}
	z\\
	y
	\end{array}} \right] = \left[ {\begin{array}{*{20}{c}}
	A&B\\
	C&D
	\end{array}} \right]\left[ {\begin{array}{*{20}{c}}
	w\\
	u
	\end{array}} \right].
\end{equation}
Then, the open loop model $P$ of the impedance control structure can be denoted as:
\begin{equation}
\left[ {\begin{array}{*{20}{c}}
	{\tilde e}\\
	{\tilde u}\\
	{{\tau _L}}\\
	e
	\end{array}} \right] 
= \left[ {\begin{array}{*{20}{c}}
	{{A_{11}}}&{{A_{12}}}&{{A_{13}}}&{{B_1}}\\
	{{A_{21}}}&{{A_{22}}}&{{A_{23}}}&{{B_2}}\\
	{{C_{11}}}&{{C_{12}}}&{{C_{13}}}&{{D_1}}\\
	{{C_{21}}}&{{C_{22}}}&{{C_{23}}}&{{D_2}}
	\end{array}} \right]
	\left[ {\begin{array}{*{20}{c}}
	{{\varphi _L}}\\
	d\\
	n\\
	{{\omega _d}}
	\end{array}} \right],
\end{equation}
where:
\begin{equation}
	\left\{
		\begin{aligned}
			& {A_{11}} = -({P_d} + {G_2}){W_\varphi }{W_e}, \quad
			{A_{12}} =  - {G_1}{W_d}{W_e}, \quad
			{A_{13}} = {G_2}{W_n}{W_e}, \quad
			{B_1} =  - {G_1}{W_e}, \\
			& {A_{21}}=0, \quad 
			{A_{22}} = 0, \quad
			{A_{23}} = 0, \quad
			{B_2} = {W_u}, \\
			& {C_{11}} = {G_2}{W_\varphi }, \quad
			{C_{12}} = {G_1}{W_d}, \quad
			{C_{13}} =  - {G_2}{W_n}, \quad
			{D_1} = {G_1}, \\
			& {C_{21}} = -({P_d} + {G_2}){W_\varphi }, \quad
			{C_{22}} =  - {G_1}{W_d}, \quad
			{C_{23}} = {G_2}{W_n},\quad 
			{D_2} = -{G_1}.
		\end{aligned}
	\right.
\end{equation}

A state space representation for the general open loop model $P$ takes the form:
\begin{equation}
	\left\{
		\begin{aligned}
			{{\dot x}_p} &= {A_p}{x_p} + {B_u}u + {B_w}w \\
			y &= {C_y}{x_p} + {D_{yw}}w \\
			z &= {C_z}{x_p} + {D_{zu}}u + {D_{zw}}w
		\end{aligned}
	\right.
\end{equation}

\section{Controller design with mixed $H_2/H_{\infty}$ synthesis}
\label{section3}

\subsection{Constraints for mixed $H_2/H_{\infty}$ synthesis}

A dynamic output feedback controller $K(s)$ is synthesized with the form:

\begin{equation}
	\left\{ 
		\begin{aligned}
		{{\dot x}_k} &= {A_k}{x_k} + {B_k}y\\
		u &= {C_k}{x_k} + {D_k}y
		\end{aligned}
 	\right..
\end{equation}

Combining the open loop system $P$ and the dynamic output feedback controller $K$, the closed loop of the impedance control system is derived as:
\begin{equation}
\left\{ 
	\begin{aligned}
	{{\dot x}_{cl}} &= \bar A{x_{cl}} + \bar Bw\\
	y &= {{\bar C}_y}{x_{cl}} + {{\bar D}_y}w\\
	z &= {{\bar C}_z}{x_{cl}} + {{\bar D}_z}w
	\end{aligned}
\right.,
\end{equation}
where:
\begin{equation}
	\begin{aligned}
		&\bar A = \left[ {\begin{array}{*{20}{c}}
						{{A_p} + {B_p}{D_k}{C_y}}&{{B_p}{C_k}}\\
						{{B_k}{C_y}}&{{A_k}}
						\end{array}} \right], \quad
		 \bar B = \left[ {\begin{array}{*{20}{c}}
						{{B_w} + {B_p}{D_k}{D_{yw}}}\\
						{{B_k}{D_{yw}}}
						\end{array}} \right], \\
		&{\bar C_y} = \left[ {\begin{array}{*{20}{c}}
							{{C_y}}&0
							\end{array}} \right], \quad
		 {\bar D_y} = {D_{yw}}, \\
		&{\bar C_z} = \left[ {\begin{array}{*{20}{c}}
							{{C_z} + {D_{zu}}{D_k}{C_y}}&{{D_{zu}}{C_k}}
							\end{array}} \right], \quad
		 {\bar D_z} = {D_{zw}} + {D_{zu}}{D_k}{D_{yw}}, \\
		&{x_{cl}} = \left[ {\begin{array}{*{20}{c}}
							{{x_p}}\\
							{{x_k}}
							\end{array}} \right].
	\end{aligned}
\end{equation}

Design of the optimal mixed $H_2/H_{\infty}$ controller in an analytical way is usually challenging, and a suboptimal controller is often obtained~\citep{ZhouKemin1998}. One formulation as in~\citep{Scherer1995}, is to find a stabilizing suboptimal controller $K(s)$ such that 
\begin{equation}\label{eq_Hinf}
{\left\| {{T_{w\tilde e}}} \right\|_\infty } \le {\gamma _e},
\end{equation}
\begin{equation}\label{eq_H2}
{\left\| {{T_{w\tilde u}}} \right\|_2} \le {\gamma _u}.
\end{equation}
Here, $\gamma_e$ and $\gamma_u$ are two positive scalers. The $H_\infty$ norm can be used to bound the worst impedance matching error in our application, as illustrated in Fig. \ref{fig_Impedance_Control_Synthesis}. The $H_2$ norm quantifies the energy consumed by the system.

\subsection{Performance characterization and controller design}   

To design a feedback control system satisfying the physical constraints, a critical step in the mixed $H_2/H_\infty$ synthesis is the determination of the weighting functions in Fig.~\ref{fig_Impedance_Control_Synthesis}. The advantages of using weighted performance specifications are obvious in our multi-objective optimization. First, some signals usually play more important roles than others. Second, these signals may not be measured in same units. Therefore, weighting functions are essential to make these signals comparable within the same impedance control structure~\citep{ZhouKemin1998}.  Third, we might be primarily interested in specific signals in a certain frequency band, such as the tracking error at the low frequency band. Therefore, frequency-dependent weighting functions must be chosen to characterize the performance.

\begin{figure}[!h]
	\centering
	\includegraphics[width=0.65\columnwidth]{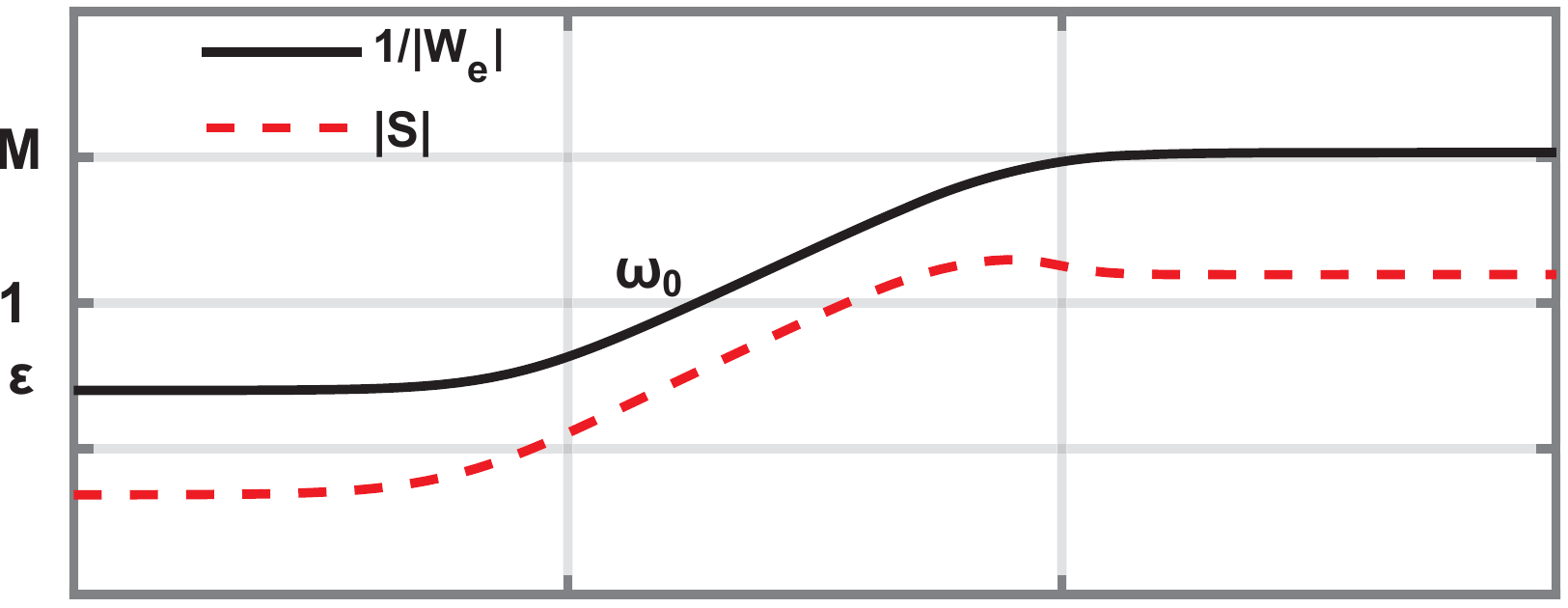}
	\caption{The weighting function $W_e(s)$ and desired sensitivity $S(s)$.}
	\label{fig_S_We}
\end{figure}

The tracking error performance of a feedback system can usually be specified in terms of the sensitivity function $S(s) = 1/(1 + P(s)K(s))$ such that:
\begin{equation}
\left| {S(j\omega )} \right| 
\le 
\left\{ {\begin{array}{*{20}{l}}
	{\varepsilon ,\;\forall \omega  \le {\omega _0}}\\
	{M,\;\forall \omega  > {\omega _0}}
	\end{array}} \right.
\end{equation}
where $\varepsilon>0$ is the desired steady tracking error, $M>0$ is the desired peak sensitivity, and $\omega_0$ is the desired bandwidth~\citep{ZhouKemin1998}. To minimize the tracking error $e$, an appropriate weighting function can be designed to satisfy:
\begin{equation}
\left| {{W_e}(j\omega )S(j\omega )} \right| \le 1
\end{equation}
with:
\begin{equation}
\left| {{W_e}(j\omega )} \right| = \left\{ {\begin{array}{*{20}{l}}
	{1/\varepsilon ,\;\forall \omega  \le {\omega _0}\;}\\
	{1/M,\;\forall \omega  > {\omega _0}}
	\end{array}} \right..
\end{equation}
Thus, $W_e(s)$ can be taken as:
\begin{equation}
{W_e}(s) = \frac{{s/M + {\omega _0}}}{{s + {\omega _0}\varepsilon }}.
\end{equation}
Then, a possible relationship between $W_e(s)$ and $S(s)$ is shown in Fig.~\ref{fig_S_We}.

To generate a control input $u$ that does not go extreme, the weighting function $W_u$ can be designed in such a way to penalize it. Taking into consideration the motor saturation, it can be set as the reciprocal of the velocity threshold. Remaining weighting functions $W_d$ and $W_n$ can be selected according to their relative magnitudes in the practical system or treated as tuning parameters by the designer to balance competing objectives of robustness and tracking performance. The weighting function $W_{\varphi}$ is designed to ensure that the general system in Fig.~\ref{fig_Impedance_Control_Synthesis} is proper, so that the mixed $H_2/H_\infty$ synthesis for rendering mass-damper-spring impedance model are valid.

\section{Results}\label{section4}

The parameters of the cable-driven SEA platform in Fig.~\ref{fig_Velocity_Source_CSEA} and weighting functions for controller parameterization in Fig.~\ref{fig_Impedance_Control_Synthesis} are listed in Table~\ref{table1}, which have been obtained by system identification experiments. Here, $r$ is the radius of the motor output shaft, and $K_g$ is the ratio of the gear head. The motor velocity threshold is 44 rad/s. In the stiffness impedance case, the weighting function $W_\varphi$ can be set as 1. In the general impedance control case, it can take a two-order low pass filter ${W_\varphi } = \left( 500/(s + 500) \right)^2$ with respect to the human movement frequency so that it brings little performance deterioration to the impedance rendering at specified low frequency band. After these parameters, the weighting functions, and the $H_2/H_\infty$ norm bounds are given, a dynamic output feedback impedance controller can be synthesized by solving the mixed $H_2/H_\infty$ suboptimal control problem under the constraints in equations~(\ref{eq_Hinf}) and (\ref{eq_H2}). Then, it can be applied to the cable-driven SEA platform to produce interactive behavior that approximates the desired impedance model.

\begin{table}[!h]
	\caption{Parameters and weighting functions for our cable-driven SEA platform.}
	\label{table1}
	\begin{center}
		\renewcommand{\arraystretch}{1.5}
		\begin{tabular}{lc|lc}
			\hline \hline
			$J_A$ & $6.90 \times {10^{ - 4}}{\rm{kg}} \times {{\rm{m}}^{\rm{2}}}$ & $b_f$ & ${\rm{0}}.{\rm{0059Nm/}}({\rm{rad/s}})$\\
			$r$ & 7.25mm & $K_g$ & 14:1\\
			$K_s$ & $2 \times 0.0242{\rm{Nm/rad}}$ & $K_{pv}$ & ${\rm{0}}{\rm{.0457Nm/(rad/s)}}$\\
			$K_{iv}$ & ${\rm{1}}{\rm{.3455Nm/(rad/s)}}$ & $M$ & 1\\
			$\omega_0$ & ${\rm{60rad/s}}$ & $\varepsilon$ & 0.005\\
			$W_u$ & $1/44$ & $W_d$ & 0.1\\
			$W_n$ & $0.1$ & $W_\varphi$ & 1 or ${\left( {\frac{{500}}{{s + 500}}} \right)^2}$\\
			\hline \hline
		\end{tabular}
	\end{center}
\end{table}

\subsection{Simulation and results}

In this work, three cases for different stiffness impedance control have been examined, with the controller synthesized with respective $H_2/H_\infty$ norm bounds. In each case, the human hand motion $\varphi_L$ is specified as a sinusoidal signal with the frequency of 2 Hz and magnitude of 2 rad. If the interaction torque $\tau_L$ tracks the desired torque $\tau_d$ well, accurate impedance control is achieved. The closed loop system is simulated by MATLAB/Simulink. 

\begin{figure}[!h]
	\centering
	\subfigure[]{\includegraphics[width=0.45\columnwidth]{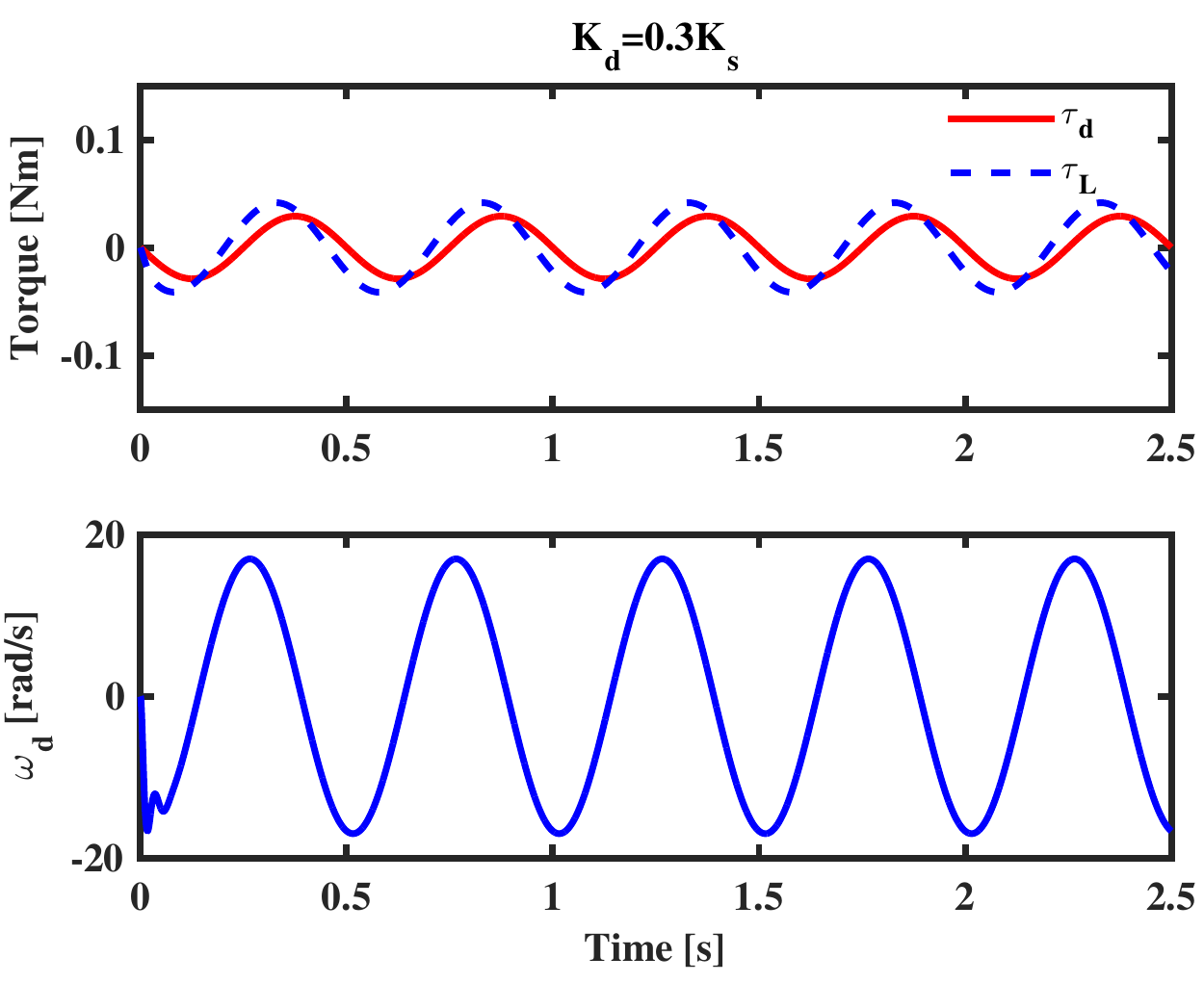}}
	\quad\quad
	\subfigure[]{\includegraphics[width=0.45\columnwidth]{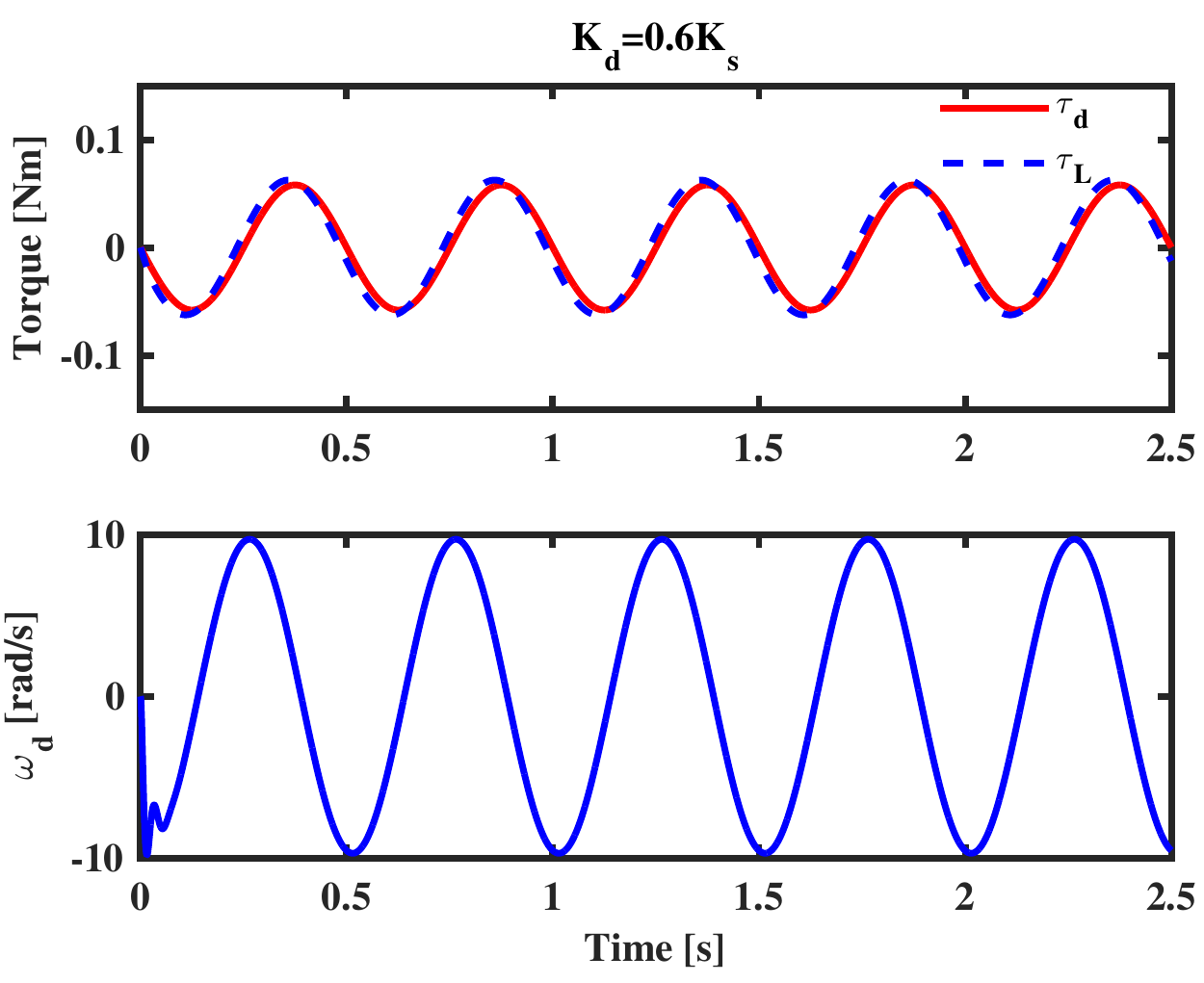}}
	\\
	\subfigure[]{\includegraphics[width=0.45\columnwidth]{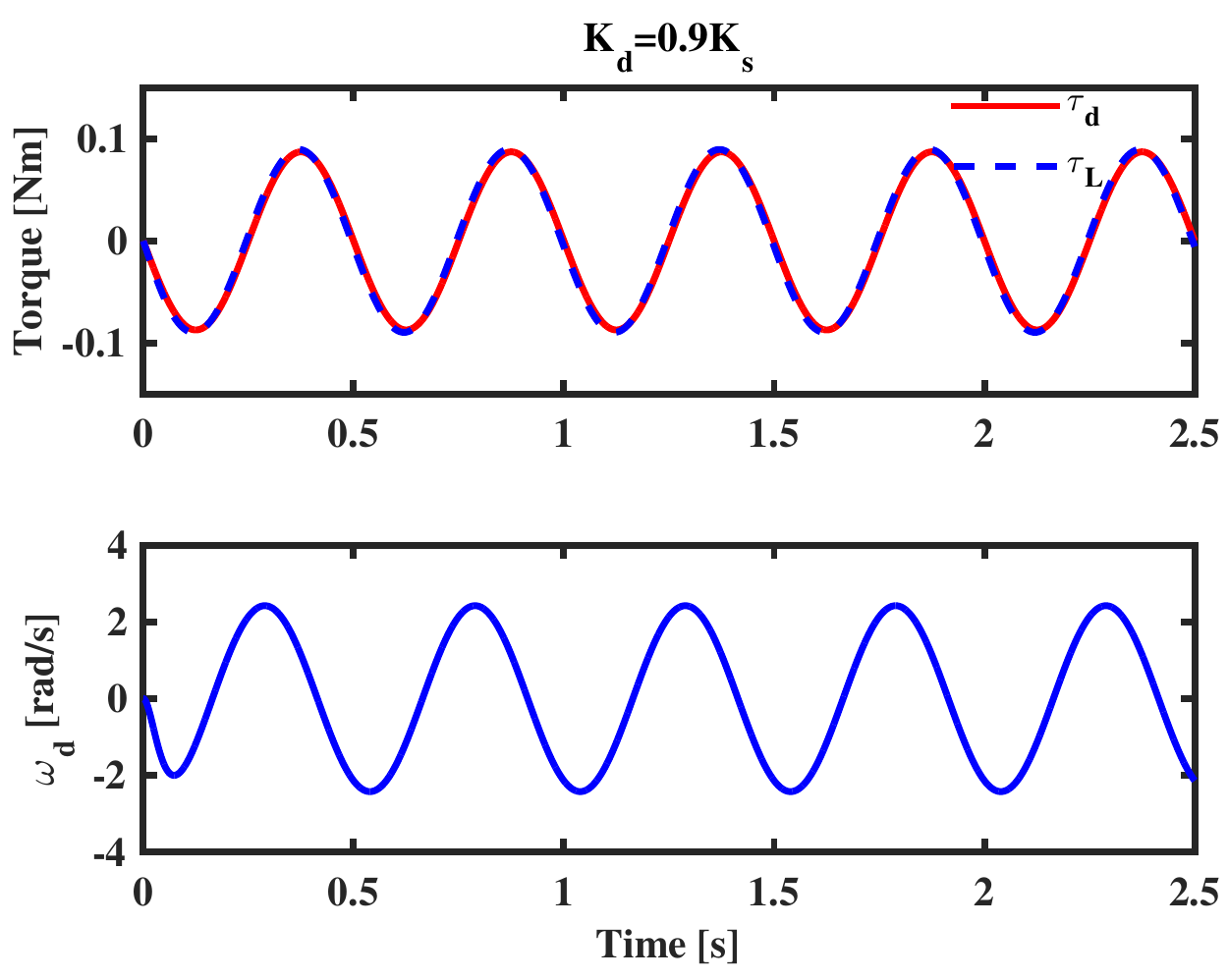}}
	\caption{Simulation results for different stiffness control.}\label{fig_simulation}
\end{figure}

The first simulation is to render the stiffness $K_d=0.3K_s$. The bounds are given as $\gamma_e=0.0580$ and $\gamma_u=43.4$. The simulation results of the desired and actual interaction torque, and the desired motor velocity are illustrated in Fig.~\ref{fig_simulation}-a. The max tracking error is 0.0233 Nm, and the max control input is 16.9859 rad/s below the saturation velocity. In the second simulation, the desired stiffness is increased to $0.6K_s$ as shown in Fig.~\ref{fig_simulation}-b. The bounds of the $H_\infty$ and $H_2$ norm are decreased to $\gamma_e=0.0330$ and $\gamma_u=29.9$. The maximal desired interaction torque is $0.0580$ Nm. The worst tracking error is $0.0130$ Nm and the maximal control input is $9.8110$ rad/s. Compared with the first simulation, the tracking error is smaller and less control effort is required. The desired stiffness is further increased to $0.9K_s$. The tracking error is even smaller and less control effort is demanded compared with the previous two simulations, as can be seen in Fig.~\ref{fig_simulation}-c. It gives the bounds of the $H_\infty$ and $H_2$ norm as $\gamma_e=0.0222$ and $\gamma_u=0.685$. The maximal desired interaction torque is 0.0870 Nm, the worst tracking error is only 0.0060 Nm, and the required maximal control input is 2.4232 rad/s.

To accurately render a smaller stiffness close to zero impedance, i.e., to achieve high transparency, higher control effort and quicker velocity response from the motor are required for the cable-driven SEA to respond to the human motion. But for a practical system, motor performance limitations may be the major restrictions for the rendering accuracy. In another special case of rendering a stiffness that equals to the physical stiffness of the elastic component, the system consumes no extra energy for torque tracking theoretically if the disturbance $d$ and noise $n$ are not taken into consideration.

\begin{figure}[!h]
	\centering
	\subfigure[]{\includegraphics[width=0.45\columnwidth]{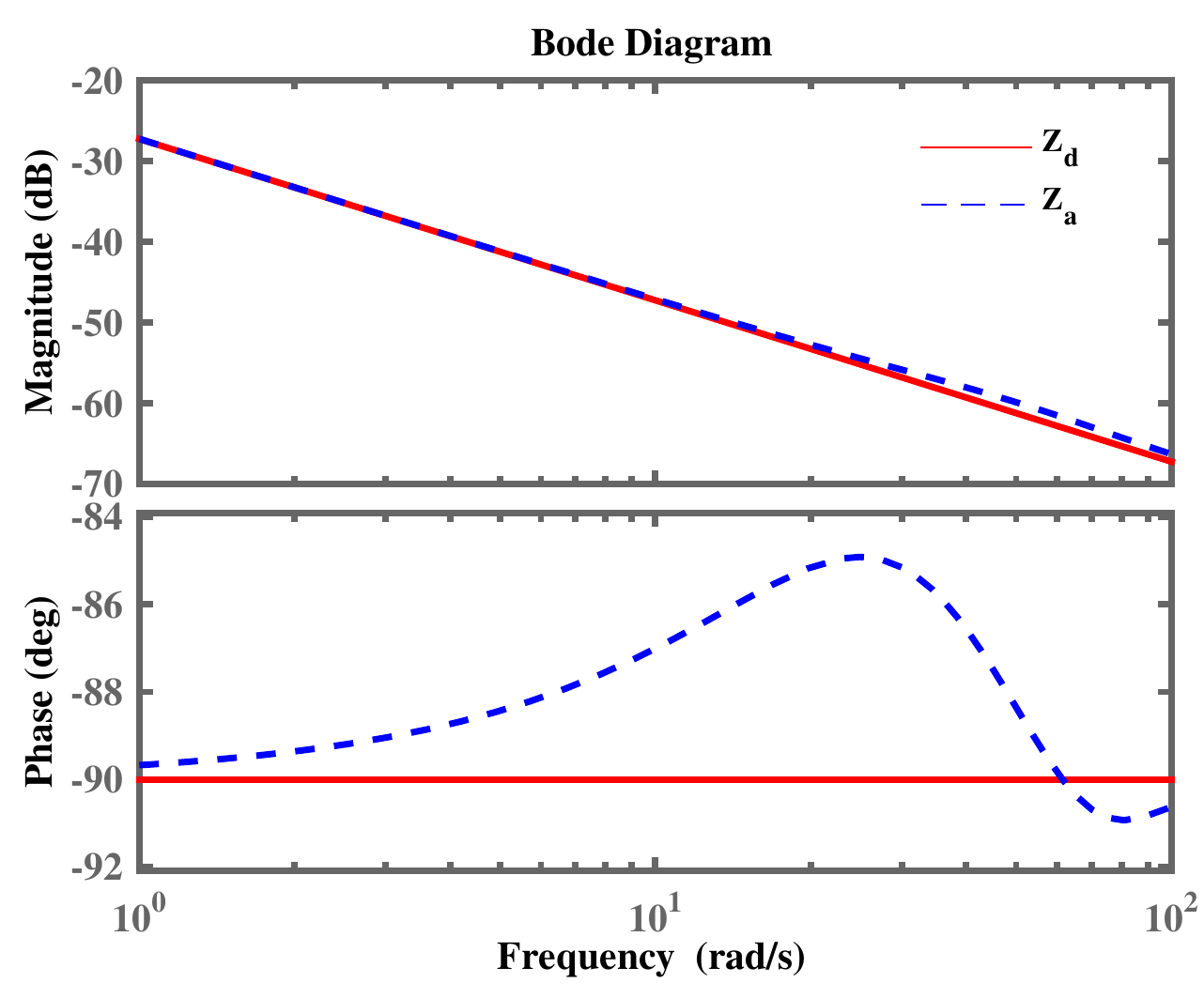}}
	\quad\quad
	\subfigure[]{\includegraphics[width=0.45\columnwidth]{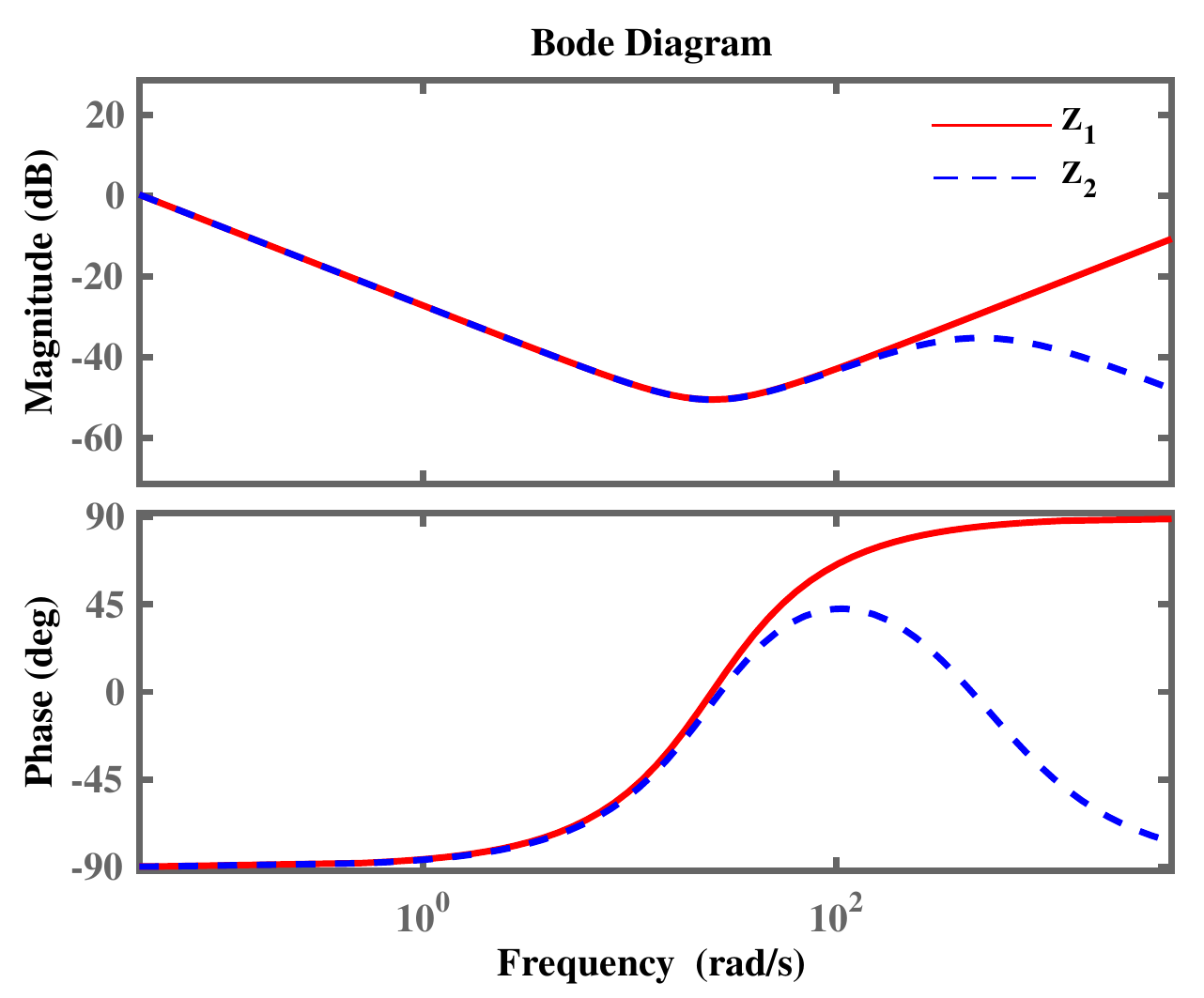}}
	\caption{(a): Bode plots of the desired impedance $Z_d(s)$ and actual impedance $Z_a(s)$ for stiffness control with $K_d=0.9K_s$. (b): The influence of the low pass filter $W_\varphi(s)$ on the impedance $Z_d(s)$ and passivity with $Z_1(s)=Z_d(s)$, $Z_2(s)=W_\varphi(s)Z_d(s)$, $K_d=0.9K_s$, $B_d=0.5b_f$ and $M_d=0.1J_A$.}
	\label{fig_bode}
\end{figure}

\subsection{Passivity analysis}

For a system to stably interact with the human arm, several constraints for $Z(s)$ defined by equation~(\ref{eq_Z}) have to be satisfied to guarantee passivity. The first constraint is to demand all closed loop poles be strictly confined to the left-half complex $s$-plane. Another is that the real part of the driving point impedance $Z(s)$ must be positive. These two constraints are also equivalent to bounding the phase of $Z(s)$ to the range of $\,\left[ { - {{90}^ \circ },\;{{90}^ \circ }} \right]$ across the entire frequency band. However, this constraint is really conservative and excessively restrains the performance. For human arm motion, the movement frequency can be limited to low frequencies, e.g., below 5 Hz. Then, the system only needs to guarantee relaxed passivity, in the sense that the phase of $Z(s)$ only needs to be in the range of $\,\left[ { - {{90}^ \circ },\;{{90}^ \circ }} \right]$ for the low frequency band.

As plotted in Fig.~\ref{fig_bode}-a, the bode plots of impedance for $K_d=0.9K_s$ indicate that the system is passive at low frequencies. Actually, for the desired stiffness below or equal to the physical stiffness $K_s$ of the elastic component in the SEA, relaxed passivity can be achieved by our impedance control strategy. Besides, less discrepancy between the magnitudes of the actual and the desired impedance indicates good rendering accuracy.

To analyze the influence of the low pass filter $W_\varphi(s)$ on the impedance and passivity, the bode plots of $Z_1(s)=Z_d(s)$ and $Z_2(s)=W_\varphi(s)Z_d(s)$ also have been drawn in Fig.~\ref{fig_bode}-b. As it can be seen, the magnitude and phase of impedance had little deviation at the specified low frequency range. Thus, the low pass filter $W_\varphi(s)$ does not deteriorate the impedance rendering performance.

\subsection{Experiment and results}

Once the impedance controller is synthesized and tested in simulation, it can be applied to our physical SEA system in Fig.~\ref{fig_prototype} for experimental verification. In this work, several experiments have been conducted with our impedance control strategy based on mixed $H_2/H_\infty$ synthesis. For each case, the human hand interacted with the handle, sliding along the guide, and the motor provides corresponding velocity and motion to regulate the interaction impedance under our designed controllers.

\begin{figure}[!h]\centering
	\begin{minipage}{\columnwidth}
		\centering
		\subfigure[]{\includegraphics[width=0.45\columnwidth]{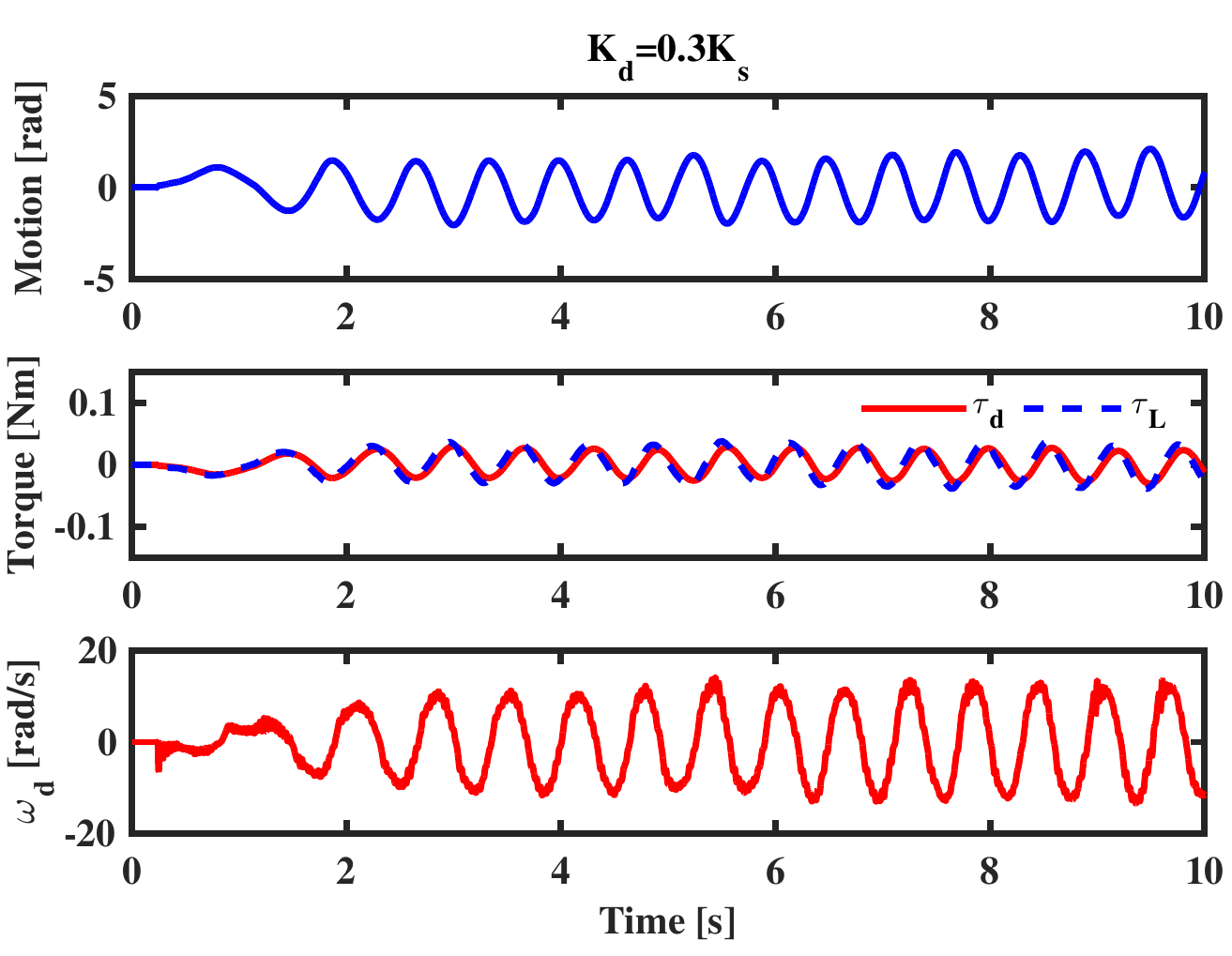}}
		\quad\quad
		\subfigure[]{\includegraphics[width=0.45\columnwidth]{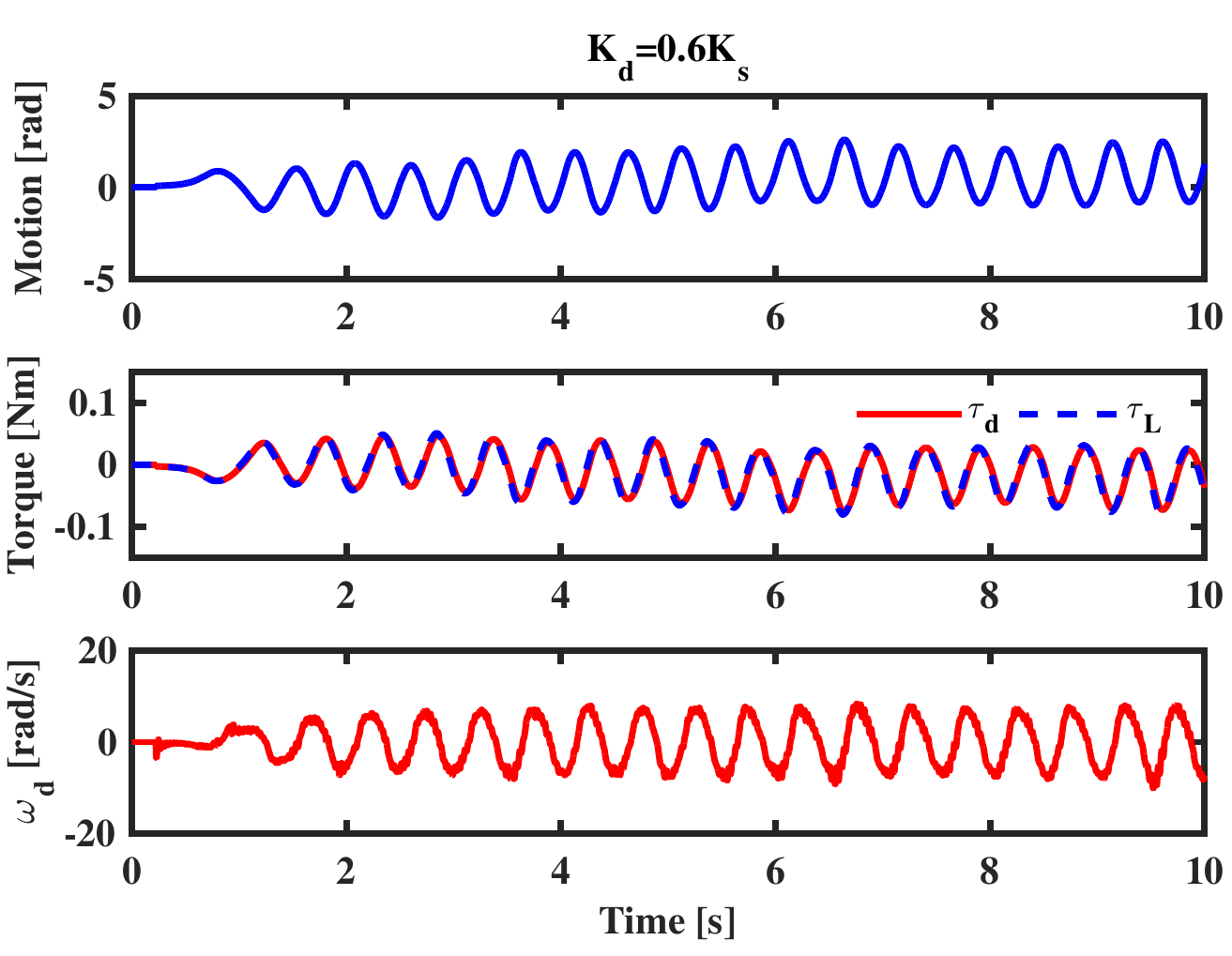}}
		\\
		\subfigure[]{\includegraphics[width=0.45\columnwidth]{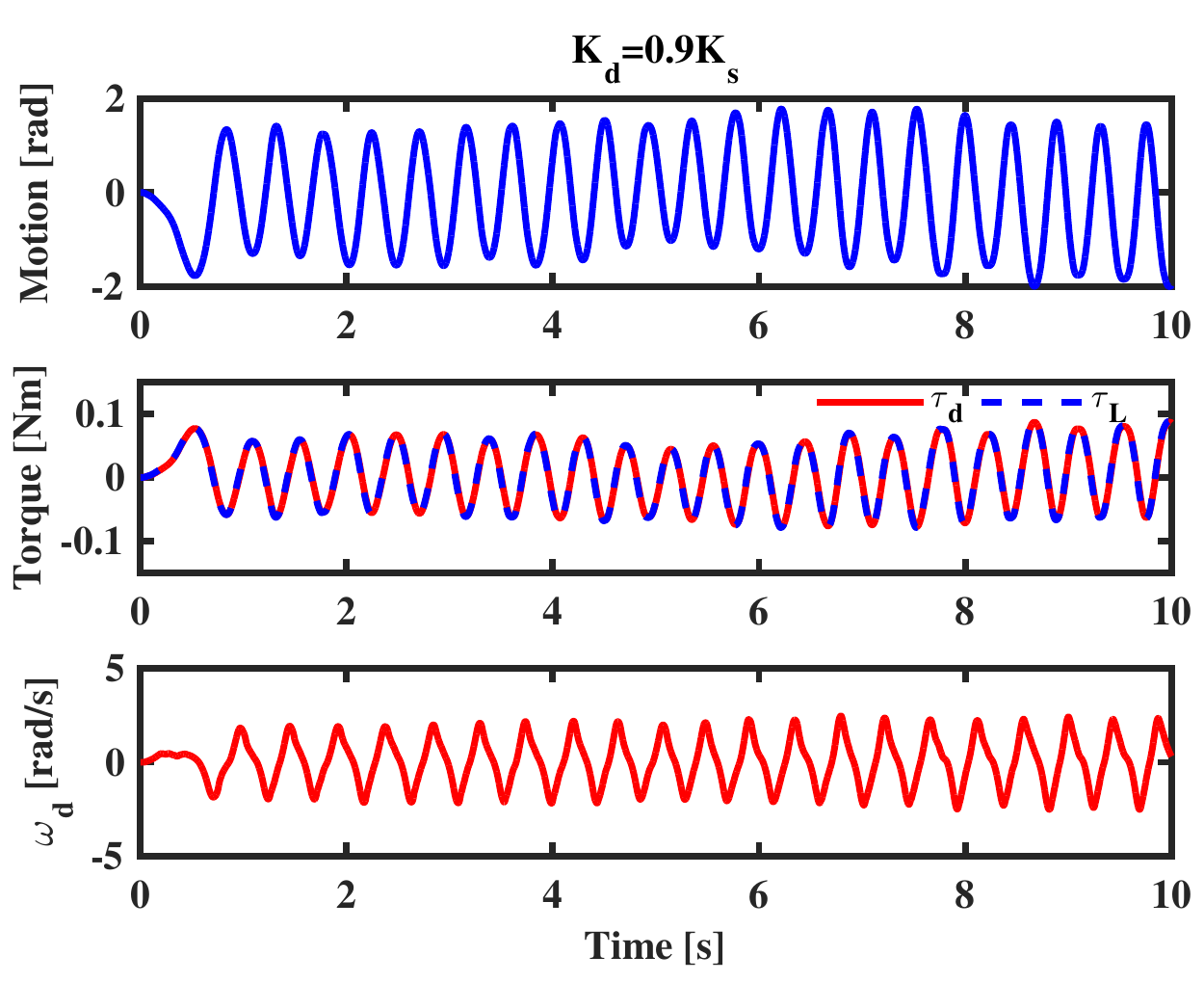}}
		\quad\quad
		\subfigure[]{\includegraphics[width=0.45\columnwidth]{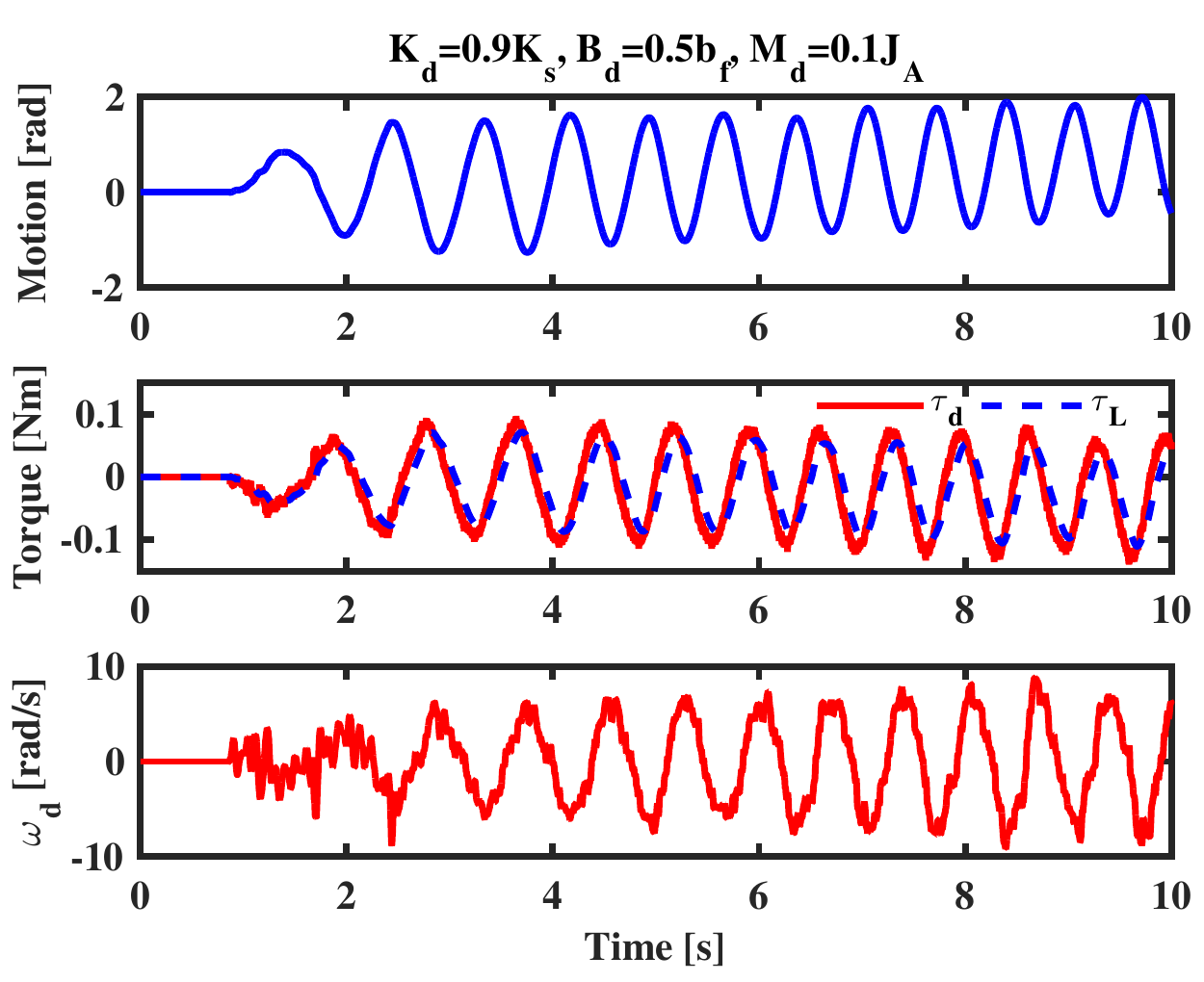}}
	\end{minipage}
	\caption{Experimental results for different desired impedance.}\label{fig_experiment}
\end{figure}

In the first experiment, stiffness rendering with $K_d=0.3K_s$ was tested. The hand motion $\varphi_L$, the desired interaction torque $\tau_d$, the actual interaction torque $\tau_L$ and the desired motor velocity $\omega_d$ are illustrated in Fig.~\ref{fig_experiment}-a. The maximal interaction torque, maximal velocity and worst tracking error are 0.0565 Nm, 14.5953 rad/s and  0.0180 Nm respectively. In the second experiment, the desired stiffness was specified as $K_d=0.6K_s$. The results are shown in Fig.~\ref{fig_experiment}-b. The maximal interaction torque, maximal velocity and worst tracking error are 0.0915 Nm, 10.1295 rad/s and 0.0128 Nm respectively. In the third experiment, the desired stiffness was increased to $K_d=0.9K_s$. The results are illustrated in Fig.~\ref{fig_experiment}-c. The maximal interaction torque, maximal velocity and worst tracking error are 0.1346 Nm, 3.1241 rad/s and 0.0079 Nm respectively.

During the last experiment, the impedance with $K_d=0.9K_s$, $B_d=0.5b_f$ and $M_d=0.1J_A$ was rendered and the results are presented in Fig.~\ref{fig_experiment}-d. The maximal interaction torque, maximal velocity and worst tracking error are 0.1153 Nm, 13.0338 rad/s and 0.0727 Nm respectively.

In the stiffness impedance control mode, the control inputs in the three experiments are all bounded within the motor's velocity limit and do not show any extreme variations. The system can hold stable interaction with human hand with guaranteed relaxed passivity. With increased stiffness, the tracking error and control effort become smaller by choosing proper $H_2/H_\infty$ norm bounds. In the general impedance rendering case, the control input and interaction torque show little oscillations due to the differentials introduced by the inertia and damping terms.

\section{Conclusion}\label{section5}

This paper has presented an impedance control method with mixed $H_2/H_\infty$ synthesis and relaxed passivity for a cable-driven SEA system to be applied for pHRI. 

The impedance control of the cable-driven SEA was reformulated into a general impedance matching framework, in which the system's impedance dynamics is shaped for matching with a predefined impedance model. The weighting functions for each signal were designed to balance competing performance requirements and to make the framework satisfying physical constraints. Then, the system structure was represented by a state space model. Constraints for the tracking error and control input were transformed to $H_\infty$ and $H_2$ norm bound with respect to the state space model. The passivity requirement for stable human-robot interaction was relaxed in such a way that the phase of the rendered impedance was confined in the range of $\,\left[ { - {{90}^ \circ },\;{{90}^ \circ }} \right]$ only for the low frequency band. Thus, the solution can be less conservative. A dynamic output feedback controller can then be synthesized by solving the mixed $H_2/H_\infty$ suboptimal control problem under the given norm bounds. Both simulation and experimental results for various desired impedance models have shown that the output torque tracked well the desired torque, no excessive vibrations happened, and the demanded motor velocities were well regulated below the motor's velocity limit. These indicated good impedance rendering performance. 

Therefore, our impedance control strategy with mixed $H_2/H_\infty$ synthesis and relaxed passivity on the low frequency band has shown good efficacy and brought significant advantages for the control of the cable-driven SEA in pHRI applications.

\section*{Acknowledgments}

The authors would like to thank the responsible reviewers and editor for the careful reviews of the paper and the constructive comments, which have been of great help to improve the quality of the paper. This work was supported by the National Natural Science Foundation of China (61403215), the Natural Science Foundation of Tianjin (13JCYBJC36600) and	the Fundamental Research Funds for the Central Universities.

\end{document}